\documentclass{article} \usepackage{iclr2027_conference,times}

\usepackage{amsmath,amsfonts,bm}

\def\eqref#1{equation~\ref{#1}}

\def\1{\bm{1}}

\DeclareMathAlphabet{\mathsfit}{\encodingdefault}{\sfdefault}{m}{sl}
\SetMathAlphabet{\mathsfit}{bold}{\encodingdefault}{\sfdefault}{bx}{n}

\usepackage{hyperref}
\usepackage{url}

\usepackage{amsmath,amsfonts,amssymb}
\usepackage{array}
\usepackage{graphicx}
\usepackage{capt-of}

\usepackage{multirow}
\usepackage{booktabs}   
\newcommand{\best}[1]{\textbf{#1}}
\newcommand{\second}[1]{\underline{#1}}
\iclrfinalcopy

\title{OmniGlimpse: Asymmetric Identity Memory for Online Multi-View Tracking}

\vspace{-20pt} 
\author{
{\bfseries
Rangya Zhang$^{1*}$,
Jiaping Xiao$^{1*}$,
Yujing Shang$^{1*}$,
Yuhang Zhang$^{1}$,\\[-1pt]
Mingsheng Li$^{1}$,
Mir Feroskhan$^{1\dagger}$
}\\[5pt]
{\normalfont\small
$^{1}$Nanyang Technological University
}
}

\begin{document}

\maketitle
\begingroup
\renewcommand{\thefootnote}{\fnsymbol{footnote}}
\footnotetext[1]{Equal contribution.}
\footnotetext[2]{Corresponding author.}
\endgroup

\begin{abstract}
Multi-view tracking extends visual coverage beyond a single camera,
providing complementary observations when targets are occluded or
leave the view. This raises a question: how can observations from
different views be matched to the correct identities without
offline scene-specific training or camera calibration?
Progress toward this goal remains limited because many offline
formulations assume such deployment-specific knowledge, while online
adaptation can amplify association errors.
We study this problem as online multi-view identity maintenance (OMIM):
targets are initialized once per view, with no further identity labels
provided after each initialization.
To this end, we introduce OmniGlimpse, an asymmetric propose--verify
framework that separates adaptive association from stable identity
consolidation. At the representation level, short-term local and
cross-view evidence is mapped into target-specific adaptive spaces to
propose associations, whereas long-term identity distributions are maintained for verification
in a canonical space defined by a frozen encoder. At the update level,
short-term evidence may guide current associations, whereas, after bootstrapping, only
verified observations may update persistent memory or drive
adaptation.
Under a unified target-specified evaluation protocol, OmniGlimpse
improves over the best evaluated online baseline for each metric by 9.1 IDF1 and 22.1 CVIDF1
points on M3Track. The corresponding gains on MvMHAT are 20.3 and
26.2 points. Despite using no offline scene-specific training, it outperforms
offline-trained references in IDF1 and remains competitive in cross-view
association. These results support an approach to online identity
maintenance that adapts to changing appearances while consolidating
reliable evidence into stable identity anchors.

\end{abstract}

\vspace{-8pt}
\noindent\begin{minipage}{0.95\textwidth}
\centering
\includegraphics[width=\linewidth]{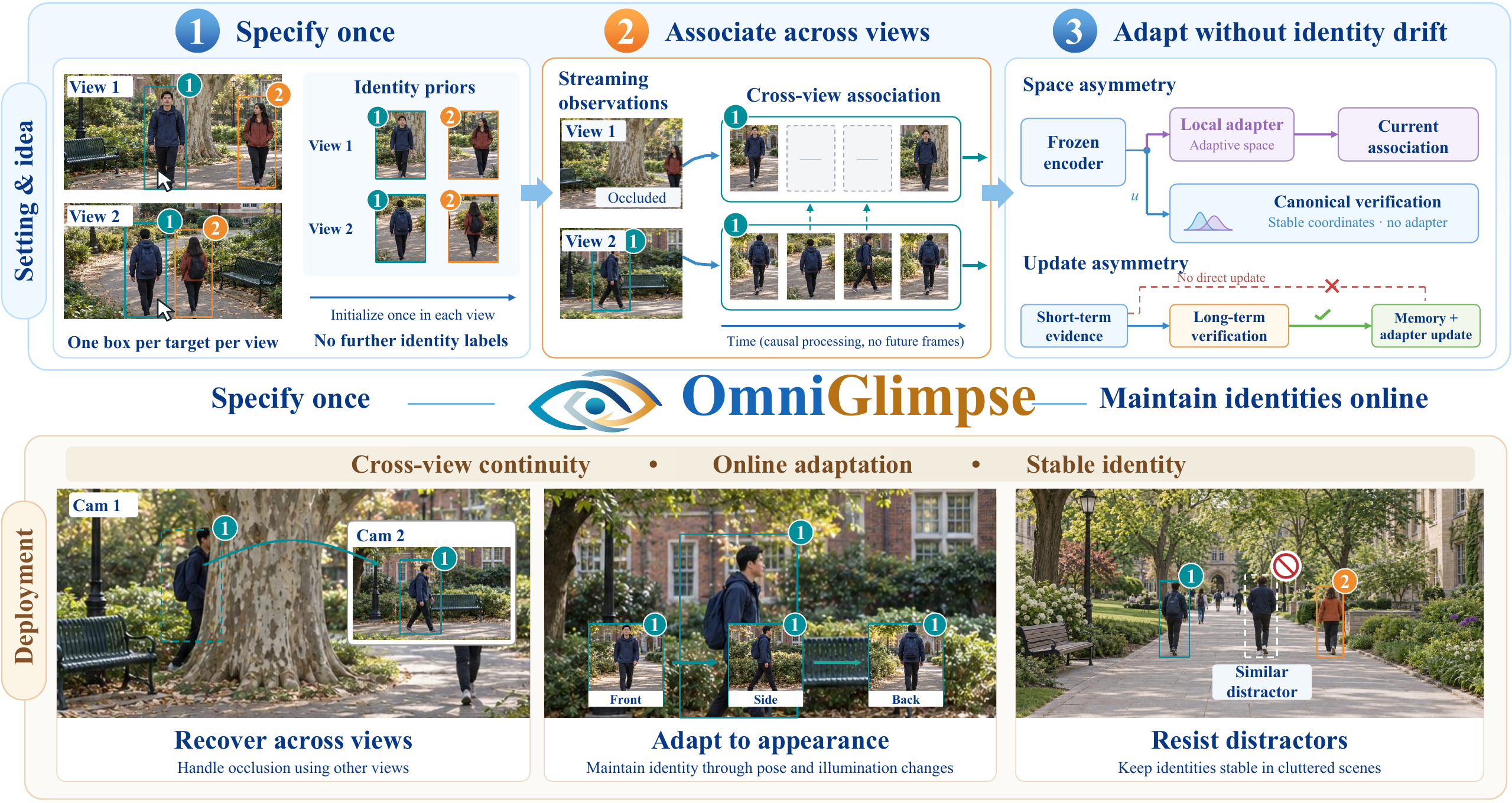}
\vspace{-20pt}
\captionof{figure}{\textbf{Overview of OmniGlimpse.} \textit{Top:} targets are initialized once per view and maintained through adaptive association, canonical verification, and verified updates. \textit{Bottom:} OmniGlimpse supports cross-view recovery, appearance adaptation, and resistance to distractors.}
\label{fig:teaser}
\end{minipage}
\clearpage

\section{INTRODUCTION}

Multi-view tracking maintains identity continuity beyond a single camera and supports long-horizon perception in surveillance~\cite{xie2024robust}, sports analysis~\cite{cui2023sportsmot}, and traffic understanding~\cite{liu2024trafficmot}. Multiple views provide complementary observations of the targets but also introduce new ambiguity in identity matching. The tracker must distinguish each target from visually similar distractors despite large viewpoint and appearance changes. Multi-view trackers associate observations through appearance, trajectories, or geometry~\cite{gan2021self,cheng2023rest}, with online variants updating tracks and appearance memories as observations arrive~\cite{xie2024robust,specker2024ocmctrack}. A related line of target-specific online methods adapts the representation to changes in a designated target's appearance~\cite{ye2022following,ye2024oclreid}.

Deploying these capabilities together in unseen multi-view scenes exposes three connected challenges.
\textbf{(i) Deployment requirements.} Multi-view multi-target (MVMT) trackers often learn appearance models offline~\cite{gan2021self,fan2025all}, while geometry-based approaches additionally require calibration or world coordinates~\cite{specker2024ocmctrack}, restricting deployment when these geometric priors are unavailable.
\textbf{(ii) Uncertain cross-view evidence.} Cross-view observations can compensate for unreliable local evidence, but view-dependent appearance gaps and asymmetric observation noise~\cite{amosa2023multi} make these cues themselves uncertain.
\textbf{(iii) Self-reinforcing errors.} When recent identity assignments are reused for online adaptation, a wrong pseudo-label can enter memory, alter the representation, and bias subsequent matches.
Together, these challenges make the problem more than online cross-view association: the tracker must also determine which observations are reliable enough to change its persistent identity state.

We refer to this as online multi-view identity maintenance (OMIM). Each target is initialized with
a bounding box at its first valid appearance in each view. Its identity is then maintained across
views without further identity labels. To address this, we introduce OMNIGLIMPSE, an asymmetric propose–verify framework that separates adaptive association from stable identity consolidation. Its design is asymmetric in both update permission and representation space. First, short-term
evidence can only propose or guide associations. After initialization and bootstrapping, only observations consistent with the long-term identity distributions may update persistent memory and
target-specific adaptive spaces. Second, short-term local and cross-view evidence is matched in
the adapters’ adaptive spaces to propose associations, while per-target identity distributions serve
as anchors for verification in a canonical space defined by a frozen encoder. This asymmetric design enables OmniGlimpse to adapt to changing appearances while preserving stable verification
and supervision. Experiments on M3Track ~\cite{fan2025all} and MvMHAT ~\cite{gan2021self} show
improved identity maintenance over online baselines, with ablations supporting the design.

Our work makes three main contributions:
\begin{itemize}
    \item We introduce OmniGlimpse for online multi-view identity
    maintenance, where targets are initialized once per view and maintained across
    views without further identity labels. The framework combines
    cross-view evidence with target-specific online adaptation
    to track changing appearances without
    offline scene-specific training or camera calibration.

    \item We develop an asymmetric propose--verify design with
    update permissions and representation spaces. Short-term evidence
    proposes associations. After bootstrapping, long-term verification controls which
    observations may update persistent memory and adaptive representations.
    A frozen canonical space anchors the evolving target-specific adapters.

    \item We evaluate OmniGlimpse on M3Track and MvMHAT under a unified
    target-specified protocol. It outperforms evaluated online
    baselines in IDF1 and CVIDF1 and remains competitive with
    offline-trained references. Ablations support the benefits of online
    adapter learning and verification-gated updates within the complete system.
\end{itemize}

\section{Related Work}

\paragraph{Single-view online identity tracking.}
Online MOT maintains identities by associating detections
with motion and appearance cues~\cite{maggiolino2023deep,shim2025focusing}. ByteTrack~\cite{zhang2022bytetrack}
exploits low-confidence detections to recover fragmented trajectories,
while OC-SORT~\cite{cao2023observation} improves motion-based association
under occlusion and non-linear motion. Template updating~\cite{zhou2025spd}
and test-time adaptation~\cite{zaveri2025improving,segu2023darth}
address changes in appearance or deployment conditions.
Target-specific methods such as Following
Closely~\cite{ye2022following} and OCLReID~\cite{ye2024oclreid} update
the target appearance model. OCLReID further uses
short-term and long-term experiences to optimize the feature extractor
online. Such adaptation improves responsiveness, but recent identity assignments
also become supervision for subsequent updates. An incorrect assignment
can therefore alter the appearance model used for future matching,
allowing a temporary association error to persist through later updates.

\paragraph{Multi-view identity association.}
Multi-view tracking extends identity maintenance by using observations from cameras when local evidence becomes unreliable.
One approach establishes cross-view correspondence from appearance and trajectory cues~\cite{huang2023enhancing,yamane2025mvtrajecter}.
MvMHAT~\cite{gan2021self} considers tracking and cross-view association, while ReST~\cite{cheng2023rest} models spatial-temporal relations through graph-based association.
Benchmarks MTMMC~\cite{woo2024mtmmc} and M3Track~\cite{fan2025all} extend evaluation to realistic scenes.
When appearance is ambiguous, MUTR3D~\cite{zhang2022mutr3d}, ADA-Track~\cite{ding2024ada}, and OCMCTrack~\cite{specker2024ocmctrack} exploit 3D queries, geometric cues, or world coordinates for cross-view matching.
Although OCMCTrack performs association online, such methods may still depend on offline appearance learning~\cite{gan2021self,fan2025all} or calibration and world-coordinate information~\cite{specker2024ocmctrack}, limiting deployment when these priors are unavailable.

\paragraph{Memory and Adaptation in Identity Tracking.}
Memory has been used in tracking and ReID systems to support identity
decisions~\cite{videnovic2026distractor}. Templates,
tracklet states, or memories are maintained to recover targets
after occlusion and stabilize appearance matching~\cite{aydemir2025track}.
Person ReID stores identity features or prototypes,
providing more stable references than the
observation alone~\cite{dai2022cluster,cheng2023crossmodality}.
Online adaptation uses unlabeled observations to update
caches or refine representations~\cite{karmanov2024efficient,segu2023darth}.
However, evidence useful for matching may not be reliable
enough to supervise updates, while representation changes can
make stored features inconsistent with observations.
OmniGlimpse addresses these issues through an asymmetric propose--verify
design that combines adaptive association with canonical-space verification
to govern persistent memory updates and learning after bootstrapping.

\section{Method}

\subsection{Problem Setup and Overview}\label{sec:problem}\label{sec:overview}

Our objective is to maintain the identities of a small set of targets of interest $\mathcal{T}=\{1,\ldots,M\}$ across multiple views $\mathcal{V}=\{1,\ldots,V\}$. Each target $m\in\mathcal{T}$ is initialized with a bounding box at its first valid appearance in each view. The target reference images $\mathcal{P}_m$ are crops extracted from these initialization boxes and become available only when the corresponding initialization occurs. No further frame-level identity annotations are provided after initialization in each view.

At time $t$, view $v$ produces an observation set $\mathcal{O}_t^{(v)}$ containing $n_t^{(v)}$ observations. Using only observations and initialization information available up to $t$, the system must infer, for every target $m$ and view $v$, the observation currently corresponding to that target:
\begin{equation}
\mathcal{O}_{t}^{(v)}=\bigl\{o_{t,i}^{(v)}\bigr\}_{i=1}^{n_t^{(v)}},
\qquad
a_t^{(m,v)}\in\mathcal{O}_{t}^{(v)}\cup\{\varnothing\}.
\end{equation}
Here, $\varnothing$ denotes that the target is absent, occluded, or cannot be identified with sufficient confidence. The objective is to maintain the identity consistency of each target across views and over time.

\begin{figure*}[t]
    \centering
    \includegraphics[width=\textwidth]{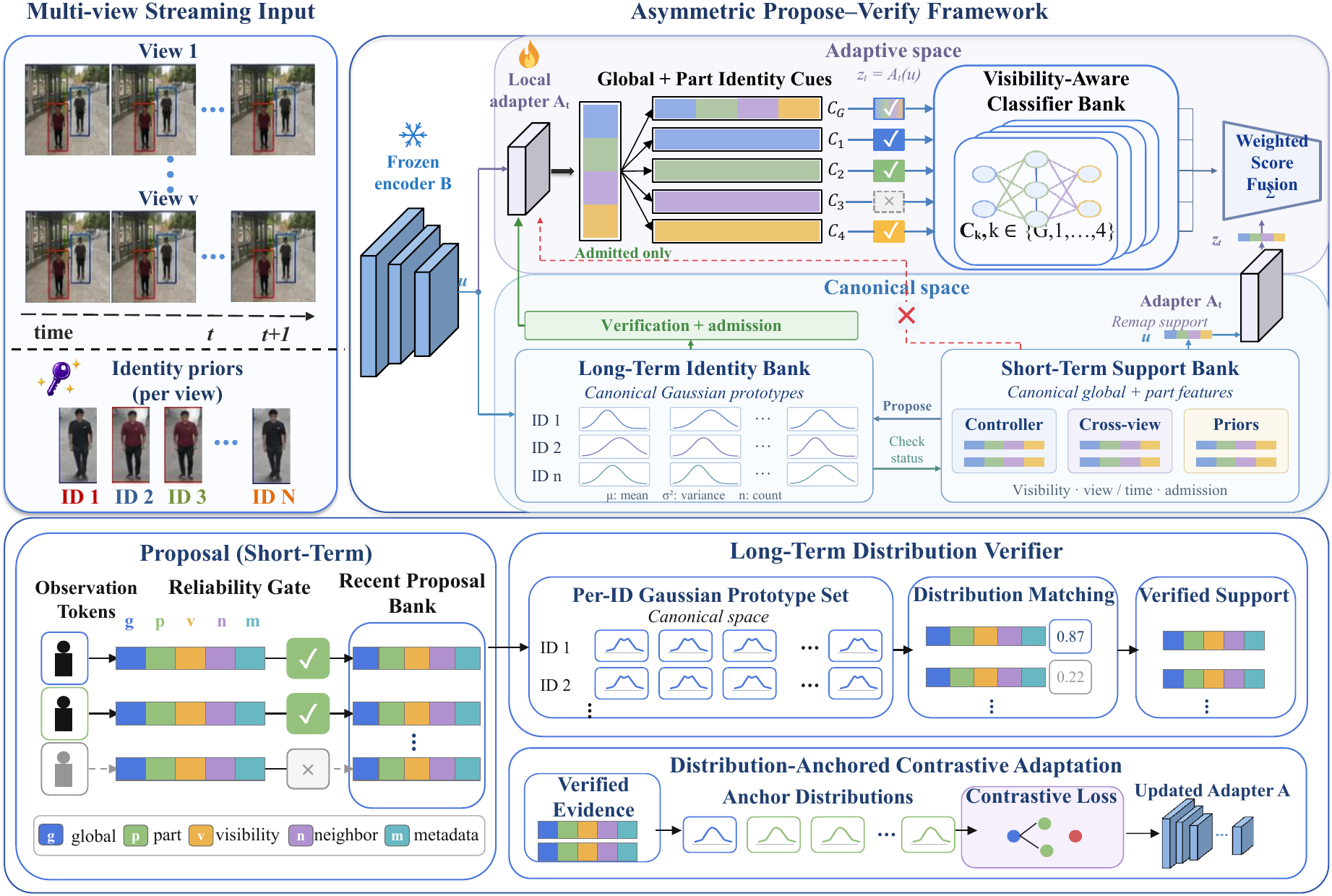}
    \caption{Overview of OmniGlimpse. Short-term evidence supports association in adaptive spaces, while long-term distributions provide canonical-space verification. After initialization and bootstrapping, only admitted observations may update persistent memory and local adapters.}
    \label{fig:overview}
\end{figure*}

Given observations and target priors, OmniGlimpse maintains target identities through a closed-loop process of target-specific association, long-term verification, and online adaptation (Fig.~\ref{fig:overview}).

\subsection{Association and Verification}\label{sec:controller}

\paragraph{Canonical and adaptive features.}
Online association must adapt to appearance changes, while cross-view evidence and long-term identity statistics need a consistent feature space across updates. At each time step, an initial detector--tracker produces person observations, each consisting of a bounding box, a person crop, and a local track identity. Following the part-based representation design in OCLReID~\cite{ye2024oclreid}, a ReID encoder $B$ extracts a global embedding $g_0(o)$ and $K$ part embeddings $\{g_k(o)\}_{k=1}^{K}$ from each observation $o$. All views use the same frozen encoder, while each target--view pair maintains a local adapter $A_t^{(m,v)}$:
\begin{equation}
u(o)=B(o)=\bigl(g_0(o),\ldots,g_K(o)\bigr),
\qquad
z_t^{(m,v)}(o)=A_t^{(m,v)}\bigl(u(o)\bigr).
\label{eq:dual_features}
\end{equation}
Here, $u(o)$ denotes the canonical features and $z_t^{(m,v)}(o)$ the adaptive features, with $z_{t,k}^{(m,v)}(o)$ denoting the global or part cue $k$. The part features are accompanied by visibility scores $w_k(o)\in[0,1]$, and an orientation head predicts $\omega(o)\in\{\mathrm{front},\mathrm{back}\}$ for orientation-aware scoring.

Memory banks store canonical features, and views exchange these features rather than local adapter outputs. For short-term association, both local and received cross-view features are mapped through the current local adapter. Prior references and long-term identity distributions instead use the canonical features for matching and verification. Online learning updates the local adapters without changing $B$, preserving the coordinate system used for communication and long-term verification. The identity distributions can still accumulate verified evidence within this fixed space.

\paragraph{Candidate association.}\label{sec:assign}
For each target--view pair $(m,v)$, candidate observations are evaluated using initialization references and adaptive ridge classifiers. The prior score measures global and part-level similarity in the canonical space, while the classifiers evaluate adaptive features under current appearance conditions. Prior evidence supports initialization and recovery; once the classifiers are available, their confidence provides the local association score $s_{\mathrm{local}}^{(m,v)}(o)$.

The active support bank contains local target samples, temporally eligible cross-view samples, and initialization references. All support features are mapped through the current local adapter before comparison. The memory score $s_{\mathrm{mem}}^{(m,v)}(o)$ aggregates top-$L$ similarities, using part evidence when it meets a discriminativeness criterion and global evidence otherwise. Cross-view evidence therefore contributes through the support bank rather than a separate additive score.

Memory assistance is activated when the current binding is missing, uncertain, or contested. For these candidates, the association score combines local confidence and memory support:
\begin{equation}
s_{\mathrm{assoc}}^{(m,v)}(o)
=
(1-\lambda_t^{(m,v)})s_{\mathrm{local}}^{(m,v)}(o)
+\lambda_t^{(m,v)}s_{\mathrm{mem}}^{(m,v)}(o).
\label{eq:association_score}
\end{equation}
Both scores are bounded in $[0,1]$, and $\lambda_t^{(m,v)}=\lambda_{\mathrm{mem}}\in[0,1]$ when assistance is active and support is available; otherwise $\lambda_t^{(m,v)}=0$. These bounded scores are not calibrated identity probabilities.

For targets in the view, we first retain uncontested bindings that satisfy the tracking conditions. For the remaining targets, eligible target--observation pairs are ranked by $s_{\mathrm{assoc}}$ and assigned greedily, allowing each target and observation to participate in at most one pair. Candidates below the acceptance threshold are discarded, and overlapping assignments are suppressed. The associations remain proposals for long-term verification; selecting a candidate does not authorize memory consolidation or learning. Scoring, support eligibility, and conflict handling are detailed in Appendix~\ref{app:association}.

\paragraph{Long-term verification.}\label{sec:memory}
The long-term memory verifies candidate associations by maintaining a consolidated identity distribution for each target. Since target appearance can be multi-modal under viewpoint changes, occlusion, and partial visibility, we represent this distribution with a small set of Gaussian prototypes. Each view $v$ maintains a local target distribution
\begin{equation}
\mathcal{G}_t^{(m,v)}
=
\bigl\{(\mu_j^{(m,v)},\sigma_j^{2,(m,v)},n_j^{(m,v)})\bigr\}_{j=1}^{P_t^{(m,v)}},
\qquad P_t^{(m,v)}\le P,
\label{eq:long_memory}
\end{equation}
where $\mu_j$ is the prototype mean, $\sigma_j^2$ its diagonal variance, and $n_j$ its admitted-sample count. These distributions use normalized canonical global features and can incorporate evidence from other views. They share a feature coordinate system, not necessarily identical distribution parameters.

For a candidate observation $o$, let $\bar u(o)=g_0(o)/\|g_0(o)\|_2$. The verifier measures its agreement with the consolidated distribution using a dimension-normalized squared Mahalanobis distance:
\begin{equation}
d_j^{(m,v)}(\bar u)
=
\frac{1}{D}\sum_{r=1}^{D}
\frac{(\bar u_r-\mu_{j,r}^{(m,v)})^2}
{\sigma_{j,r}^{2,(m,v)}+\epsilon},
\qquad
\ell_{\mathrm{long}}^{(m,v)}(o)
=
\exp\!\left[-\frac12\min_j d_j^{(m,v)}(\bar u(o))\right].
\label{eq:mahalanobis}
\end{equation}
Here, $D$ is the global feature dimension and $\epsilon>0$ stabilizes the denominator. The diagonal variance accounts for uncertainty along different feature dimensions, while the minimum selects the best-matching appearance mode. The resulting consistency score is not a normalized identity probability.

Verification uses the distribution available before admitting the current observation. Once the count $n_t^{(m,v)}=\sum_j n_j^{(m,v)}$ reaches $N_{\min}$, a candidate passes if $\ell_{\mathrm{long}}^{(m,v)}(o)\ge\tau_\ell$ and fails otherwise. Before this point, the result is unavailable, neither failed nor verified. Initialization references seed the distribution, which matures through conservative bootstrap admission (Appendix~\ref{app:long_memory}).

A failed check prevents a new or recovered association from being committed. An established binding may remain an output if it still satisfies the tracking conditions, but the failed observation cannot update persistent memory or train the adapters or classifiers. Passing verification indicates consistency with the identity model; further admission rules govern updates. This separates evidence for current association from evidence allowed to modify persistent identity state.

\subsection{Verification-Gated Updates}\label{sec:updates}

Association outputs and learning samples have different admission requirements. After assignment, an output observation may enter the short-term support bank to assist subsequent proposals, but its presence there does not authorize training or long-term consolidation. Each stored observation retains its admission status. Persistent identity memory and positive training buffers accept only evidence that satisfies the update rule below, apart from initialization and restricted bootstrap.

For a local observation $o$, let $V_t^{(m,v)}(o)=1$ only when the mature long-term verifier returns pass, and zero when it returns fail or unavailable. A positive observation is admitted if
\begin{equation}
\begin{aligned}
\mathcal{A}_{+,t}^{(m,v)}(o)=\mathbb{I}\bigl[
&\,o=a_t^{(m,v)} \ \land\ V_t^{(m,v)}(o)=1\\
&{}\land\ s_{\mathrm{assoc}}^{(m,v)}(o)\ge\tau_a\\
&{}\land\ \min\{\Delta_{\mathrm{tgt}}(o),\Delta_{\mathrm{obs}}(o)\}
    \ge\tau_\Delta\\
&{}\land\ T_{\mathrm{stable}}^{(m,v)}(o)\ge\tau_T
\bigr].
\end{aligned}
\label{eq:contrastive_pos_admission}
\end{equation}
Here, $a_t^{(m,v)}$ is the final retained assignment, the two margins are computed before allocation, and $T_{\mathrm{stable}}$ counts consecutive observed frames with the same target--track binding. Missing observations or a changed binding reset this count. The rule controls long-memory admission and positive samples for both Ridge fitting and adapter learning; it is not a universal condition for emitting an association. Initialization labels provide initial supervision. Before maturity, conservative bootstrap can populate the long-term distribution but cannot authorize ordinary pseudo-label training. Remote evidence requires independent receiver-side admission (Appendix~\ref{app:long_memory}).

All admission decisions are made before any memory or parameter update. We first fix the admitted sample set using the current assignments and pre-update distributions, then use that set for long-memory maintenance and online learning. The Gaussian parameters used as adaptation anchors are cached before these updates, so an observation cannot modify its own anchor within the same step. The updated distributions are used for verification only at the next step.

\subsection{Distribution-Anchored Adaptation}\label{sec:contrastive_adaptation}

The propose--verify memory maintains candidate-level identity consistency, but appearance changes can still weaken feature discriminability. We therefore refine the local adapters using admitted memory samples. Each adapter applies a residual transformation to the canonical cues,
\begin{equation}
z_{t,k}^{(m,v)}(o)=g_k(o)+h_{\phi_{t,k}^{(m,v)}}\bigl(g_k(o)\bigr),
\qquad
\widehat z_{t,k}^{(m,v)}(o)=
\frac{z_{t,k}^{(m,v)}(o)}{\|z_{t,k}^{(m,v)}(o)\|_2}.
\label{eq:residual_adapter}
\end{equation}
The residual branches initially output zero, so the normalized adaptive features start from their canonical counterparts. The frozen encoder defines the coordinates, while residual branches learn target-specific corrections. Architecture and sampling details are given in Appendix~\ref{app:adaptation}.

For each target--view pair, the memory system constructs same-view and cross-view triplets $(a,p,n)$ from admitted evidence. The anchor $a$ and positive $p$ represent the same target; negatives come from observations admitted for other specified targets. Unassigned detections and unresolved overlapping observations are excluded rather than assumed to be reliable negatives. All three samples are mapped through the same current local adapter, including those received from other views. Same-view triplets preserve local discrimination, while cross-view positives bring different appearances together within the receiver's adaptive space. For jointly visible cues, we use the loss
\begin{equation}
\ell_{\mathrm{tri},k}(a,p,n)=
\bigl[
\cos(\widehat z_k(a),\widehat z_k(n))
-\cos(\widehat z_k(a),\widehat z_k(p))
+\delta_{\mathrm{tri}}
\bigr]_+.
\label{eq:triplet_loss}
\end{equation}
Here, target, view, and time indices are omitted for readability. Averaging over available cues and triplets gives $\mathcal L_{\mathrm{sv}}$ and $\mathcal L_{\mathrm{cv}}$ for same-view and cross-view evidence.

Triplet learning separates target observations from distractors. To also regularize adaptation with accumulated identity evidence, we penalize deviations of admitted positive global features from the canonical Gaussian prototypes:
\begin{equation}
\begin{aligned}
\mathcal L_{\mathrm{anchor}}
&=\frac{1}{|\mathcal P_+|}
\sum_{o\in\mathcal P_+}
\min_j d_{j,t}^{(m,v)}\bigl(\widehat z_{t,0}^{(m,v)}(o)\bigr),\\
\mathcal L_{\mathrm{adapt}}
&=\mathcal L_{\mathrm{sv}}+\gamma_{\mathrm{cv}}\mathcal L_{\mathrm{cv}}
+\beta_{\mathrm{dist}}\mathcal L_{\mathrm{anchor}}.
\end{aligned}
\label{eq:adapt_loss}
\end{equation}
Here, $\mathcal P_+$ contains the positive observations in the training batch, and $d_{j,t}^{(m,v)}$ uses the cached pre-update Gaussian parameters in Eq.~\ref{eq:mahalanobis}. The residual initialization makes this penalty compare aligned coordinates at the start of adaptation; the penalty subsequently discourages departures from the canonical identity distribution. It constrains the adaptive global features, whereas inference-time verification continues to use the unmodified canonical features.

During adaptation, Gaussian parameters are detached and gradients update only the local residual branches. After an update, cached canonical features are remapped through the new adapter, and Ridge classifiers are refit using only samples with training permission. The refreshed features and classifiers support the next association step without another backbone pass on stored crops. Buffer-triggered refitting and missing-class handling are detailed in Appendix~\ref{app:association}.

\section{Experiments}
\label{sec:experiments}

Our experiments address two questions:
(1) Can OmniGlimpse maintain target identities across views without identity labels after initialization?
(2) Can identity verification support online adaptation while limiting the propagation of association errors?
We examine its behavior in sequences.

\begin{table*}[t]
\centering
\caption{Selected-target results on M3Track. Offline references use target-selection information, including ground-truth-assisted trajectory selection for CrossMOT and AGW and full-sequence appearance matching for MvMHAT (Appendix~\ref{app:evaluation}). Best and second-best in bold and underline.}
\label{tab:m3track_results}
\resizebox{\textwidth}{!}{
\begin{tabular}{l|ccccccc|ccccc}
\toprule
Method 
& HOTA$\uparrow$ & DetA$\uparrow$ & AssA$\uparrow$
& IDF1$\uparrow$ & MOTA$\uparrow$ & MOTP$\uparrow$ & IDs$\downarrow$
& MDA$\uparrow$ & CVIDP$\uparrow$ & CVIDR$\uparrow$
& CVIDF1$\uparrow$ & CVMA$\uparrow$\\
\midrule
\multicolumn{13}{c}{Offline-trained MVMT references} \\
\midrule
MvMHAT
& .5832 & .4407 & \best{.7723} & .6267 & .4641 & \best{.9335} & 11
& .1620 & .2914 & .1920 & .2277 & .3204\\
CrossMOT
& \second{.5970} & \second{.5540} & .6445 & \second{.7945} & \second{.6628} & \second{.8182} & 116
& \best{.5635} & \second{.6150} & \second{.5499} & \second{.5786} & \second{.6335}\\
AGW
& .4748 & .3928 & .5773 & .6347 & .4630 & .7936 & 9
& .2350 & .3374 & .2556 & .2864 & .3717\\
\midrule
\multicolumn{13}{c}{Online baselines} \\
\midrule
OCLReID
& .5862 & .4810 & \second{.7173} & .7424 & .5795 & .7972 & 18
& .3290 & .4480 & .3552 & .3918 & .4586\\
OC-SORT+RPF
& .5071 & .3904 & .6614 & .6508 & .4603 & .7960 & \best{2}
& .2678 & .3771 & .2892 & .3237 & .3984\\
ByteTrack
& .4664 & .3610 & .6073 & .5872 & .3967 & .8016 & 8
& .2582 & .3282 & .2687 & .2916 & .3607\\
BoT-SORT+ReID
& .4651 & .3559 & .6128 & .5944 & .4013 & .7980 & \second{5}
& .2892 & .4036 & .2955 & .3375 & .3889\\
\midrule
\textbf{Ours (Full)}
& \best{.6438} & \best{.5861} & .7102 & \best{.8333} & \best{.7217} & .7945 & 28
& \second{.5503} & \best{.6576} & \best{.5792} & \best{.6129} & \best{.6677}\\
\bottomrule
\end{tabular}}
\end{table*}

\begin{table*}[t]
\centering
\caption{Selected-target results on MvMHAT. Offline references use information for target selection: CrossMOT uses ground-truth-assisted trajectory selection; MvMHAT and AGW use full-sequence appearance matching (Appendix~\ref{app:evaluation}). Best and second-best results are shown in bold and underline.}
\label{tab:mvmhat_results}
\resizebox{\textwidth}{!}{
\begin{tabular}{l|ccccccc|ccccc}
\toprule
Method 
& HOTA$\uparrow$ & DetA$\uparrow$ & AssA$\uparrow$
& IDF1$\uparrow$ & MOTA$\uparrow$ & MOTP$\uparrow$ & IDs$\downarrow$
& MDA$\uparrow$ & CVIDP$\uparrow$ & CVIDR$\uparrow$
& CVIDF1$\uparrow$ & CVMA$\uparrow$\\
\midrule
\multicolumn{13}{c}{Offline-trained MVMT references} \\
\midrule
MvMHAT
& .3646 & .1786 & \second{.7443} & .3012 & .1448 & \second{.9556} & \best{1}
& .0500 & .0522 & .0505 & .0513 & .2354\\
CrossMOT
& .5217 & \second{.4448} & .6130 & \second{.6901} & \second{.4874} & .8196 & 17
& \best{.5626} & \best{.6380} & \best{.5819} & \best{.6078} & \best{.7190}\\
AGW
& \second{.5761} & .4352 & \best{.7628} & .5875 & .3518 & \best{.9993} & 6
& .3448 & .4687 & .3759 & .4131 & .5114\\
\midrule
\multicolumn{13}{c}{Online baselines} \\
\midrule
OCLReID
& .4644 & .3060 & .7133 & .5292 & .3290 & .8403 & \second{3}
& .2280 & .2975 & .2454 & .2676 & .4245\\
OC-SORT+RPF
& .3444 & .2008 & .6024 & .3984 & .2319 & .7833 & \best{1}
& .0786 & .1112 & .0868 & .0964 & .2878\\
ByteTrack
& .4516 & .3112 & .6651 & .5130 & .3104 & .8448 & 9
& .2719 & .3330 & .2875 & .3068 & .4494\\
BoT-SORT+ReID
& .4198 & .2742 & .6509 & .4661 & .2754 & .8514 & 4
& .2114 & .2764 & .2278 & .2481 & .4007\\
\midrule
\textbf{Ours (Full)}
& \best{.5820} & \best{.4863} & .7135 & \best{.7323} & \best{.5060} & .8168 & 36
& \second{.5189} & \second{.6053} & \second{.5408} & \second{.5685} & \second{.6601}\\
\bottomrule
\end{tabular}}
\end{table*}

\begin{table}[t]
\centering
\caption{Ablation study on 19 non-dark M3Track test instances. The upper block shows incremental variants and the lower block full-model controls (Appendix~\ref{app:ablation_controls}). V-Upd. denotes verification-gated updates; \textit{Adaptive Query} uses adapted features. Best and second-best are in bold and underline.}
\label{tab:ablation_m3track}
\small
\setlength{\tabcolsep}{3.4pt}
\resizebox{\textwidth}{!}{
\begin{tabular}{lccccc|ccccccccc}
\toprule
Variant & Rel. & Cross & Adapt. & Gauss. & V-Upd. & HOTA$\uparrow$ & DetA$\uparrow$ & IDF1$\uparrow$ & MOTA$\uparrow$ & IDs$\downarrow$ & MDA$\uparrow$ & CVIDP$\uparrow$ & CVIDF1$\uparrow$ & CVMA$\uparrow$ \\
\midrule
Base & -- & -- & -- & -- & -- & .6240 & .5548 & .7999 & .6801 & \best{19} & .3995 & .5602 & .4881 & .5395 \\
+ Reliability Controller & $\surd$ & -- & -- & -- & -- & .6386 & .5802 & .8246 & .6989 & 46 & .4769 & .6271 & .5608 & .5805 \\
+ Cross-view Assistance & $\surd$ & $\surd$ & -- & -- & -- & .6389 & \second{.5890} & .8280 & .7100 & 45 & \second{.5360} & \second{.6460} & \second{.6000} & .6250 \\
+ Online Contrastive & $\surd$ & $\surd$ & $\surd$ & -- & -- & \best{.6528} & \best{.5964} & \best{.8453} & \best{.7322} & 83 & .5255 & .6271 & .5850 & .6398 \\
\midrule
Full w/o Rel. & -- & $\surd$ & $\surd$ & $\surd$ & $\surd$ & .6178 & .5482 & .7938 & .6710 & 32 & .3924 & .5507 & .4796 & .5333 \\
w/o Online & $\surd$ & $\surd$ & -- & $\surd$ & $\surd$ & .6250 & .5650 & .8050 & .6900 & \second{26} & .5140 & .6210 & .5710 & .6210 \\
w/o Verified Updates & $\surd$ & $\surd$ & $\surd$ & $\surd$ & -- & .6390 & .5880 & .8240 & .7090 & 57 & .5190 & .6220 & .5830 & .6410 \\
Adaptive Query & $\surd$ & $\surd$ & $\surd$ & $\surd$ & $\surd$ & .6370 & .5810 & .8250 & .7110 & 44 & .5330 & .6410 & .5960 & \second{.6510} \\
\midrule
\textbf{Ours (Full)} & $\surd$ & $\surd$ & $\surd$ & $\surd$ & $\surd$ & \second{.6438} & .5861 & \second{.8333} & \second{.7217} & 28 & \best{.5503} & \best{.6576} & \best{.6129} & \best{.6677} \\
\bottomrule
\end{tabular}
}
\end{table}

\subsection{Experimental Setup}

\paragraph{Benchmarks and protocol.}
M3Track~\cite{fan2025all} is our main benchmark, providing synchronized RGB and infrared streams from two moving-camera views. We use only the visible-light streams and exclude extremely low-light nighttime sequences, yielding 36 single-view RGB sequences for training offline references and 38 for testing, corresponding to 19 paired-view test instances. MvMHAT~\cite{gan2021self} provides 3--4 synchronized views per video group; following its original split, we use 13 groups for training offline references and 13 for testing. In the online protocol, each target is initialized with a bounding box at its first annotated appearance in each view, with no subsequent frame-level identity labels. All methods are evaluated on the same selected identities; offline target-selection procedures are detailed in Appendix~\ref{app:evaluation}.

\paragraph{Baselines.}
We compare with online methods and offline-trained references. Online baselines are OCLReID~\cite{ye2024oclreid}, OC-SORT+RPF~\cite{cao2023observation}, ByteTrack~\cite{zhang2022bytetrack}, and BoT-SORT+ReID~\cite{aharon2022bot}. They receive initialization boxes but no scene-specific training and run independently per view without cross-view information exchange. Corresponding targets share an exported label only to normalize the identity namespace, providing no cross-view association or recovery cues. OCLReID and OC-SORT+RPF retain their target-identification mechanisms, whereas ByteTrack and BoT-SORT follow the tracker identity bound at initialization. Offline references are MvMHAT~\cite{gan2021self}, CrossMOT~\cite{hao2024divotrack}, and AGW~\cite{ye2021deep}. Since their target selection may use full-sequence information, including GT-assisted trajectory selection, we report them as references rather than identical-condition baselines.

\paragraph{Metrics.}
Single-view over-time identity maintenance is evaluated by aggregating results independently computed in each view, including HOTA, DetA, AssA, IDF1, MOTA, MOTP, and identity switches (IDs)~\cite{bernardin2008evaluating,ristani2016performance,luiten2021hota}. IDs denotes TrackEval's standard IDSW count. Cross-view identity association is evaluated using CVIDP, CVIDR, CVIDF1, MDA, and CVMA~\cite{gan2021self,fan2025all}. Initialization fixes the selected targets and output labels; metric matching follows the evaluation rules in Appendix~\ref{app:evaluation}.

\paragraph{Implementation.}
We use YOLOX~\cite{ge2021yolox} and BoT-SORT~\cite{aharon2022bot} to generate frame-level observations for each view. Following OCLReID~\cite{ye2024oclreid}, observations are encoded by a ResNet-18 based ReID backbone. Association, memory admission, and adapter fitting are detailed in Appendix~\ref{app:method_details}, with numerical settings in Appendix~\ref{app:configuration}.

\subsection{Overall Results}
\label{sec:overall_results}

\paragraph{M3Track.}

Table~\ref{tab:m3track_results} reports the single-view and cross-view metrics on the M3Track~\cite{fan2025all} benchmark. Compared with online baselines such as OCLReID, BoT-SORT, ByteTrack, and OC-SORT+RPF, our method achieves higher IDF1 and CVIDF1. In particular, it increases IDF1 from .7424 to .8333 and CVIDF1 from .3918 to .6129 over OCLReID, the strongest online baseline on both metrics. These gains indicate improved identity continuity within views and consistency across views under the reported evaluation measures. We further compare with offline-trained MVMT references. Relative to CrossMOT, our method improves IDF1 from .7945 to .8333 and CVIDF1 from .5786 to .6129, while its MDA is lower (.5503 versus .5635). AssA, MOTP, and IDs are not the best in the table. These results show competitive performance against offline references, whose additional target-selection information is detailed in Appendix~\ref{app:evaluation}.

\paragraph{MvMHAT.}
Table~\ref{tab:mvmhat_results} reports the comparison with online and offline methods on MvMHAT. The gains over online baselines extend to this benchmark: IDF1 increases from .5292 for OCLReID to .7323, while CVIDF1 increases from .3068 for ByteTrack to .5685. These are the strongest online baselines for the metrics. Compared with CrossMOT, our method achieves higher IDF1 (.7323 versus .6901), but lower CVIDF1 (.5685 versus .6078). Its cross-view metrics remain below CrossMOT, and it incurs more identity switches. Together, the benchmarks show IDF1 and CVIDF1 gains over the online baselines, while the comparison with offline references remains mixed.

\begin{figure}[!t]
\centering
\includegraphics[width=\linewidth]{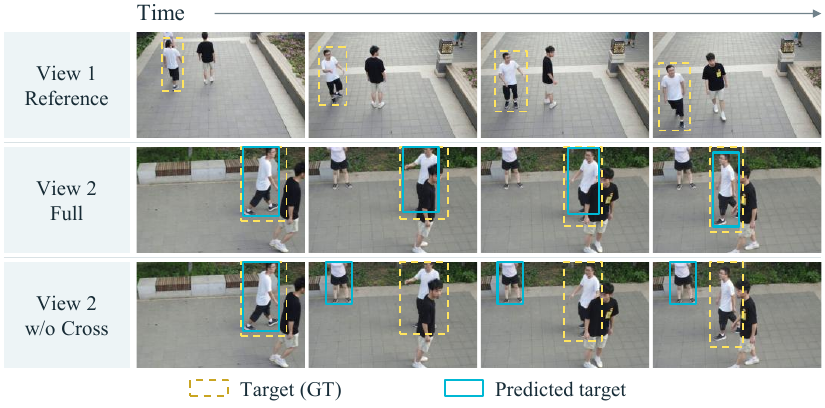}
\caption{Identity maintenance in the M3Track garden scene. Columns show synchronized observations; the first row shows the reference view. Yellow dashed boxes mark the target, and cyan boxes mark confirmed predictions.}
\label{fig:qualitative}
\end{figure}

\subsection{Ablation Study}
\label{sec:ablation}

Table~\ref{tab:ablation_m3track} examines component combinations and full-model controls (Appendix~\ref{app:ablation_controls}). Reliability control improves IDF1 and CVIDF1, while cross-view assistance raises CVIDF1 from .5608 to .6000. Adding online contrastive adaptation improves HOTA and IDF1 but lowers CVIDF1 and increases IDs. Full adds Gaussian verification with its write and learning constraints, raising CVIDF1 to .6129 and reducing IDs from 83 to 28, at the cost of lower IDF1 than the Online Contrastive variant.

The controls show that online adapter learning remains useful within the system. Disabling adapter optimization while retaining Ridge refitting lowers IDF1 and CVIDF1 by 2.83 and 4.19 points. With adapter learning retained, removing verification-gated updates lowers CVIDF1 by 2.99 points and increases IDs from 28 to 57. These comparisons support learning from new observations while restricting which observations may update persistent memory and supervise adaptation.

Using adapted rather than canonical verification queries lowers CVIDF1 by 1.69 points and increases IDs from 28 to 44. This supports retaining canonical verification queries while adapting representations for association. Because Gaussian statistics remain canonical in this control, the comparison addresses query choice under canonical statistics, not the necessity of the entire two-space design.

\subsection{Qualitative Analysis}
\label{sec:qualitative}

Figure~\ref{fig:qualitative} shows an identity maintenance example on M3Track. As the target moves near other pedestrians, the configuration without cross-view assistance assigns the target identity to another person wearing a similar white shirt. This incorrect association persists through the displayed observations, while the selected target remains visible. With cross-view assistance, the controller maintains the selected identity through the same interval. The other view provides complementary observations of the target, and the recovery step records the use of cross-view memory. This example illustrates cross-view assistance, not the isolated effects of long-term verification or online adaptation. Appendix~\ref{app:qualitative} compares incorrect assignments and interrupted outputs in the wood scene.

\section{Conclusion}
\label{sec:conclusion}

In this paper, we study online multi-view identity maintenance, where targets are specified once and tracked across views without further identity supervision. To address this problem, we propose \textsc{OmniGlimpse}, which uses asymmetric memory roles to separate association proposals from verified updates, and asymmetric representation spaces to adapt appearance features while keeping identity verification in a stable shared space. Together, these choices support tracking and adaptation across views with only initialization supervision, without scene-specific training or camera calibration. Experiments on M3Track and MvMHAT show stronger identity maintenance than the evaluated online baselines and competitive performance against offline-trained references. The current evaluation focuses on a small set of designated targets with per-view initialization; scaling the framework to denser target sets and weaker initialization remains an important direction. We will release the code to support reproducibility and future research. We hope OmniGlimpse will advance multi-view tracking toward sustained identity maintenance in changing environments.

\section*{AI Use Statement}

Generative AI tools were used to assist with language polishing and to support literature retrieval and discovery during the preparation of this manuscript. All retrieved materials were independently checked by the authors, and all AI-assisted text was reviewed and revised to ensure that it accurately reflects the authors' intended meaning and the reported work. The authors take full responsibility for the final manuscript, including its technical content, claims, citations, and conclusions.

\section*{Reproducibility Statement}

Implementation details are provided in
Appendix~\ref{app:method_details}, with model configurations
and online-update settings in Appendix~\ref{app:configuration}.
Evaluation protocols and baseline adaptations are described in
Appendix~\ref{app:evaluation}, and ablation controls are specified
in Appendix~\ref{app:ablation_controls}.
We will release the code and experiment configurations
to facilitate reproducibility.

\bibliography{iclr2027_conference}

@article{ye2024oclreid,
  author  = {Ye, H. and Zhao, J. and Zhan, Y. and Chen, W. and He, L. and Zhang, H.},
  title   = {Person re-identification for robot person following with online continual learning},
  journal = {IEEE Robotics and Automation Letters},
  year    = {2024}
}

@article{ge2021yolox,
  title={Yolox: Exceeding yolo series in 2021},
  author={Ge, Zheng and Liu, Songtao and Wang, Feng and Li, Zeming and Sun, Jian},
  journal={arXiv preprint arXiv:2107.08430},
  year={2021}
}

@article{aharon2022bot,
  title={BoT-SORT: Robust associations multi-pedestrian tracking},
  author={Aharon, Nir and Orfaig, Roy and Bobrovsky, Ben-Zion},
  journal={arXiv preprint arXiv:2206.14651},
  year={2022}
}

@inproceedings{gan2021self,
  title={Self-supervised multi-view multi-human association and tracking},
  author={Gan, Yiyang and Han, Ruize and Yin, Liqiang and Feng, Wei and Wang, Song},
  booktitle={Proceedings of the 29th ACM international conference on multimedia},
  pages={282--290},
  year={2021}
}

@inproceedings{cui2023sportsmot,
  title={Sportsmot: A large multi-object tracking dataset in multiple sports scenes},
  author={Cui, Yutao and Zeng, Chenkai and Zhao, Xiaoyu and Yang, Yichun and Wu, Gangshan and Wang, Limin},
  booktitle={Proceedings of the IEEE/CVF international conference on computer vision},
  pages={9921--9931},
  year={2023}
}

@inproceedings{woo2024mtmmc,
  title={MTMMC: a large-scale real-world multi-modal camera tracking benchmark},
  author={Woo, Sanghyun and Park, Kwanyong and Shin, Inkyu and Kim, Myungchul and Kweon, In So},
  booktitle={Proceedings of the IEEE/CVF conference on computer vision and pattern recognition},
  pages={22335--22346},
  year={2024}
}

@inproceedings{xie2024robust,
  title={A robust online multi-camera people tracking system with geometric consistency and state-aware re-id correction},
  author={Xie, Zhenyu and Ni, Zelin and Yang, Wenjie and Zhang, Yuang and Chen, Yihang and Zhang, Yang and Ma, Xiao},
  booktitle={Proceedings of the IEEE/CVF Conference on Computer Vision and Pattern Recognition},
  pages={7007--7016},
  year={2024}
}

@inproceedings{liu2024trafficmot,
  title={TrafficMOT: a challenging dataset for multi-object tracking in complex traffic scenarios},
  author={Liu, Lihao and Cheng, Yanqi and Deng, Zhongying and Wang, Shujun and Chen, Dongdong and Hu, Xiaowei and Li{\`o}, Pietro and Sch{\"o}nlieb, Carola-Bibiane and Aviles-Rivero, Angelica},
  booktitle={Proceedings of the 32nd ACM International Conference on Multimedia},
  pages={1265--1273},
  year={2024}
}

@inproceedings{cheng2023rest,
  title={Rest: A reconfigurable spatial-temporal graph model for multi-camera multi-object tracking},
  author={Cheng, Cheng-Che and Qiu, Min-Xuan and Chiang, Chen-Kuo and Lai, Shang-Hong},
  booktitle={Proceedings of the IEEE/CVF international conference on computer vision},
  pages={10051--10060},
  year={2023}
}

@inproceedings{zhang2022bytetrack,
  title={Bytetrack: Multi-object tracking by associating every detection box},
  author={Zhang, Yifu and Sun, Peize and Jiang, Yi and Yu, Dongdong and Weng, Fucheng and Yuan, Zehuan and Luo, Ping and Liu, Wenyu and Wang, Xinggang},
  booktitle={European conference on computer vision},
  pages={1--21},
  year={2022},
  organization={Springer}
}

@inproceedings{fan2025all,
  title={All-day multi-camera multi-target tracking},
  author={Fan, Huijie and Qiao, Yu and Zhen, Yihao and Zhao, Tinghui and Fan, Baojie and Wang, Qiang},
  booktitle={Proceedings of the Computer Vision and Pattern Recognition Conference},
  pages={16892--16901},
  year={2025}
}

@inproceedings{specker2024ocmctrack,
  title={OCMCTrack: Online multi-target multi-camera tracking with corrective matching cascade},
  author={Specker, Andreas},
  booktitle={Proceedings of the IEEE/CVF Conference on Computer Vision and Pattern Recognition},
  pages={7236--7244},
  year={2024}
}

@article{amosa2023multi,
  title={Multi-camera multi-object tracking: A review of current trends and future advances},
  author={Amosa, Temitope Ibrahim and Sebastian, Patrick and Izhar, Lila Iznita and Ibrahim, Oladimeji and Ayinla, Lukman Shehu and Bahashwan, Abdulrahman Abdullah and Bala, Abubakar and Samaila, Yau Alhaji},
  journal={Neurocomputing},
  volume={552},
  pages={126558},
  year={2023},
  publisher={Elsevier}
}

@inproceedings{aydemir2025track,
  title={Track-on: Transformer-based online point tracking with memory},
  author={Aydemir, G{\"o}rkay and Cai, Xiongyi and Xie, Weidi and Guney, Fatma},
  booktitle={International Conference on Learning Representations},
  volume={2025},
  pages={57112--57134},
  year={2025}
}

@inproceedings{dai2022cluster,
  title={Cluster contrast for unsupervised person re-identification},
  author={Dai, Zuozhuo and Wang, Guangyuan and Yuan, Weihao and Zhu, Siyu and Tan, Ping},
  booktitle={Proceedings of the Asian conference on computer vision},
  pages={1142--1160},
  year={2022}
}

@inproceedings{cheng2023crossmodality,
  title={Cross-Modality Person Re-identification with Memory-Based Contrastive Embedding},
  author={Cheng, De and Wang, Xiaolong and Wang, Nannan and Wang, Zhen and Wang, Xiaoyu and Gao, Xinbo},
  booktitle={Proceedings of the AAAI Conference on Artificial Intelligence},
  volume={37},
  number={1},
  pages={425--432},
  year={2023},
  doi={10.1609/aaai.v37i1.25116}
}

@article{ye2022following,
  title={Following closely: A robust monocular person following system for mobile robot},
  author={Ye, Hanjing and Zhao, Jieting and Pan, Yaling and Chen, Weinan and Zhang, Hong},
  journal={arXiv preprint arXiv:2204.10540},
  year={2022}
}

@inproceedings{cao2023observation,
  title={Observation-centric sort: Rethinking sort for robust multi-object tracking},
  author={Cao, Jinkun and Pang, Jiangmiao and Weng, Xinshuo and Khirodkar, Rawal and Kitani, Kris},
  booktitle={Proceedings of the IEEE/CVF conference on computer vision and pattern recognition},
  pages={9686--9696},
  year={2023}
}

@inproceedings{zhang2022mutr3d,
  title={Mutr3d: A multi-camera tracking framework via 3d-to-2d queries},
  author={Zhang, Tianyuan and Chen, Xuanyao and Wang, Yue and Wang, Yilun and Zhao, Hang},
  booktitle={Proceedings of the IEEE/CVF Conference on Computer Vision and Pattern Recognition},
  pages={4537--4546},
  year={2022}
}

@inproceedings{ding2024ada,
  title={Ada-track: End-to-end multi-camera 3d multi-object tracking with alternating detection and association},
  author={Ding, Shuxiao and Schneider, Lukas and Cordts, Marius and Gall, Juergen},
  booktitle={Proceedings of the IEEE/CVF Conference on Computer Vision and Pattern Recognition},
  pages={15184--15194},
  year={2024}
}

@article{hao2024divotrack,
  title={Divotrack: A novel dataset and baseline method for cross-view multi-object tracking in diverse open scenes},
  author={Hao, Shengyu and Liu, Peiyuan and Zhan, Yibing and Jin, Kaixun and Liu, Zuozhu and Song, Mingli and Hwang, Jenq-Neng and Wang, Gaoang},
  journal={International Journal of Computer Vision},
  volume={132},
  number={4},
  pages={1075--1090},
  year={2024},
  publisher={Springer}
}

@article{ye2021deep,
  title={Deep learning for person re-identification: A survey and outlook},
  author={Ye, Mang and Shen, Jianbing and Lin, Gaojie and Xiang, Tao and Shao, Ling and Hoi, Steven CH},
  journal={IEEE transactions on pattern analysis and machine intelligence},
  volume={44},
  number={6},
  pages={2872--2893},
  year={2021},
  publisher={IEEE}
}

@inproceedings{huang2023enhancing,
  title={Enhancing multi-camera people tracking with anchor-guided clustering and spatio-temporal consistency id re-assignment},
  author={Huang, Hsiang-Wei and Yang, Cheng-Yen and Jiang, Zhongyu and Kim, Pyong-Kun and Lee, Kyoungoh and Kim, Kwangju and Ramkumar, Samartha and Mullapudi, Chaitanya and Jang, In-Su and Huang, Chung-I and others},
  booktitle={Proceedings of the IEEE/CVF Conference on Computer Vision and Pattern Recognition},
  pages={5239--5249},
  year={2023}
}

@inproceedings{yamane2025mvtrajecter,
  title={MVTrajecter: Multi-View Pedestrian Tracking with Trajectory Motion Cost and Trajectory Appearance Cost},
  author={Yamane, Taiga and Masumura, Ryo and Suzuki, Satoshi and Orihashi, Shota},
  booktitle={Proceedings of the IEEE/CVF International Conference on Computer Vision},
  pages={13270--13280},
  year={2025}
}

@article{videnovic2026distractor,
  title={Distractor-Aware Memory-Based Visual Object Tracking},
  author={Videnovi{\'c}, Jovana and Kristan, Matej and Luke{\v{z}}i{\v{c}}, Alan},
  journal={International Journal of Computer Vision},
  volume={134},
  number={5},
  pages={211},
  year={2026},
  publisher={Springer}
}

@inproceedings{karmanov2024efficient,
  title={Efficient test-time adaptation of vision-language models},
  author={Karmanov, Adilbek and Guan, Dayan and Lu, Shijian and El Saddik, Abdulmotaleb and Xing, Eric},
  booktitle={Proceedings of the IEEE/CVF Conference on Computer Vision and Pattern Recognition},
  pages={14162--14171},
  year={2024}
}

@article{bernardin2008evaluating,
  title={Evaluating multiple object tracking performance: the clear mot metrics},
  author={Bernardin, Keni and Stiefelhagen, Rainer},
  journal={EURASIP Journal on Image and Video Processing},
  volume={2008},
  number={1},
  pages={246309},
  year={2008},
  publisher={Springer}
}

@inproceedings{ristani2016performance,
  title={Performance measures and a data set for multi-target, multi-camera tracking},
  author={Ristani, Ergys and Solera, Francesco and Zou, Roger and Cucchiara, Rita and Tomasi, Carlo},
  booktitle={European conference on computer vision},
  pages={17--35},
  year={2016},
  organization={Springer}
}

@article{luiten2021hota,
  title={Hota: A higher order metric for evaluating multi-object tracking},
  author={Luiten, Jonathon and Osep, Aljosa and Dendorfer, Patrick and Torr, Philip and Geiger, Andreas and Leal-Taix{\'e}, Laura and Leibe, Bastian},
  journal={International journal of computer vision},
  volume={129},
  number={2},
  pages={548--578},
  year={2021},
  publisher={Springer}
}

@article{zhou2025spd,
  title={{SPD-Updater}: Symmetric positive definite manifold geometry based temporal updating for visual object tracking},
  author={Zhou, Jinglin and Xu, Tianyang and Zhu, Xuefeng and Wu, Xiao-Jun and Kittler, Josef},
  journal={Neural Networks},
  volume={193},
  pages={107985},
  year={2026},
  doi={10.1016/j.neunet.2025.107985},
  publisher={Elsevier}
}

@inproceedings{zaveri2025improving,
  title={Improving accuracy and generalization for efficient visual tracking},
  author={Zaveri, Ram and Patel, Shivang and Gu, Yu and Doretto, Gianfranco},
  booktitle={Proceedings of the Winter Conference on Applications of Computer Vision},
  pages={9450--9460},
  year={2025}
}

@inproceedings{shim2025focusing,
  title={Focusing on tracks for online multi-object tracking},
  author={Shim, Kyujin and Ko, Kangwook and Yang, Yujin and Kim, Changick},
  booktitle={2025 IEEE/CVF Conference on Computer Vision and Pattern Recognition (CVPR)},
  pages={11687--11696},
  year={2025},
  organization={IEEE}
}

@inproceedings{segu2023darth,
  title={{DARTH}: Holistic test-time adaptation for multiple object tracking},
  author={Segu, Mattia and Schiele, Bernt and Yu, Fisher},
  booktitle={2023 IEEE/CVF International Conference on Computer Vision (ICCV)},
  pages={9683--9693},
  year={2023},
  organization={IEEE}
}

@inproceedings{maggiolino2023deep,
  title={{Deep OC-SORT}: Multi-pedestrian tracking by adaptive re-identification},
  author={Maggiolino, Gerard and Ahmad, Adnan and Cao, Jinkun and Kitani, Kris},
  booktitle={2023 IEEE International Conference on Image Processing (ICIP)},
  pages={3025--3029},
  year={2023},
  organization={IEEE}
}
\bibliographystyle{iclr2027_conference}

\appendix
\section{Method Details}\label{app:method_details}
\subsection{Candidate Association Details}\label{app:association}

\paragraph{Prior identity reference.}
While the classifiers adapt online, initialization provides an identity reference before pseudo-labels are available. For target $m$ in view $v$, let $\mathcal{P}_{m,v}^{\mathrm{init}}$ contain the crops supplied at that view's initialization. A reference is constructed for each available cue:
\begin{equation}
r_k^{(m,v)}
=
\frac{1}{|\mathcal{P}_{m,v,k}^{\mathrm{init}}|}
\sum_{p\in\mathcal{P}_{m,v,k}^{\mathrm{init}}}g_k(p),
\qquad k\in\{0,\ldots,K\},
\label{eq:init_ref}
\end{equation}
where $\mathcal{P}_{m,v,k}^{\mathrm{init}}$ includes only initialization crops in which cue $k$ is available. The global cue is always available. These canonical references remain fixed after the corresponding initialization; reference crops from another view can enter the support bank only after that view's initialization has occurred and its message has arrived. The prior score combines global cosine similarity with the mean cosine similarity over jointly available parts, using weights 0.7 and 0.3, respectively. When no part is jointly available, only the global similarity is used. The resulting score is clipped to $[0,1]$.

\paragraph{Orientation-aware classifiers.}
Each target--view unit maintains orientation-aware ridge classifiers
$\{c_{\omega,k}^{(m,v)}\}$, where $\omega\in\{\mathrm{front},\mathrm{back}\}$ and $k=0$ denotes the global cue. Each classifier is trained as a binary target/non-target regressor with labels one and zero. For observation $o$, only trained classifiers matching its predicted orientation and visible cues are evaluated. Let $\mathcal{K}_t^{(m,v)}(o)$ denote this set. Their mean response is
\begin{equation}
C_t^{(m,v)}(o)
=
\frac{1}{|\mathcal{K}_t^{(m,v)}(o)|}
\sum_{k\in\mathcal{K}_t^{(m,v)}(o)}
c_{\omega(o),k}^{(m,v)}
\bigl(\widehat z_{t,k}^{(m,v)}(o)\bigr).
\label{eq:conf}
\end{equation}
The local association score is $\operatorname{clip}(C_t^{(m,v)}(o),0,1)$. If no applicable classifier is trained, the prior score is used instead. This clipping gives a bounded ranking score, not a calibrated posterior probability. All target-specific classifiers use the same feature-normalization convention, ridge regularization, and equal total weights for positive and negative training samples. Fusion weights and acceptance thresholds are shared across targets rather than fitted independently to each target. These common settings constrain score construction but do not guarantee cross-target calibration. The classifier buffer stores canonical features and their admitted labels; it does not treat every emitted association as a training sample. Refitting uses features mapped by the current local adapter.

For each orientation/cue classifier, let $\mathcal B_+$ and $\mathcal B_-$ contain admitted positive and negative samples with the required orientation and visible cue. Negatives carry an admitted identity different from the target, as specified in Appendix~\ref{app:adaptation}. With $x_i$ the normalized feature mapped through the current target-specific adapter, fitting solves
\begin{equation}
\min_{w,b}\;
\frac{1}{2|\mathcal B_+|}\sum_{i\in\mathcal B_+}(w^\top x_i+b-1)^2
+\frac{1}{2|\mathcal B_-|}\sum_{i\in\mathcal B_-}(w^\top x_i+b)^2
+\lambda_R\|w\|_2^2,
\qquad \lambda_R>0.
\label{eq:ridge_fit}
\end{equation}
The intercept is unpenalized, and each class receives half the total sample weight. The response is $c(x)=w^\top x+b$. A classifier is fitted only when both sets are nonempty. Otherwise it is disabled and excluded from $\mathcal K_t^{(m,v)}(o)$; if that set is empty, scoring falls back to the prior. Initialization labels can supply either class, but bootstrap pseudo-labels cannot. A single-target deployment without admitted negatives therefore uses prior and memory scores rather than a one-class Ridge fit.

An affected classifier is refit whenever its admitted buffer changes, including insertion, eviction, or a change in training eligibility, even if the adapter does not change. An adapter update invalidates every affected orientation/cue model: its training features are remapped from canonical storage and it is refit before the next association step. Missing either class disables the model rather than retaining stale weights. If neither its buffer nor its adapter changes, the fitted model is reused. Newly available initialization labels are incorporated before scoring; other admitted samples affect subsequent steps only.

\paragraph{Support bank and temporal eligibility.}
The support bank combines recent local target observations, received target observations from other views, and available initialization references. Each sample retains its source view, timestamp, local track identity, canonical global/part features, and visibility mask. Remote samples are eligible only if their timestamps do not exceed the current processing time and their age is within a synchronization window $\Delta_{\mathrm{sync}}$. Features and masks are inserted and removed together. A missing part mask makes the sample unavailable for part comparison, but does not prevent global comparison. Tentative short-term support may help propose an association without being admitted for representation learning.

\paragraph{Global and part support.}
For brevity, suppress $(m,v,t)$ and let $z_k(o)$ and $z_k(b)$ denote the query and support features produced by the same current local adapter. Global pair similarity is
$h_g(o,b)=\cos(z_0(o),z_0(b))$.
For each support sample $b$, define the jointly visible parts
$\mathcal{J}(o,b)=\{k\in\{1,\ldots,K\}:w_k(o)>0,\ w_k(b)>0\}$ and
\begin{equation}
h_p(o,b)
=
\frac{1}{|\mathcal{J}(o,b)|}
\sum_{k\in\mathcal{J}(o,b)}
\cos\bigl(z_k(o),z_k(b)\bigr).
\end{equation}
Samples with $\mathcal{J}(o,b)=\varnothing$ are excluded from part aggregation. The global and part support scores, $G(o)$ and $P(o)$, are the means of the largest $L$ valid pair similarities in their respective channels. Fewer than $L$ samples use all available values; an empty channel returns zero.

Part scores are used where available when a discriminativeness criterion is met: among all current observations for the target, at least two have positive part support, the largest part score is at least $\tau_p$, and its margin over the second largest is at least $\delta_p$. For candidates with positive part support under this condition, $s_{\mathrm{mem}}(o)=\operatorname{clip}(P(o),0,1)$; otherwise it is $\operatorname{clip}(G(o),0,1)$. Thus missing or ambiguous part evidence falls back to global support. An empty support bank disables memory assistance rather than treating a zero memory score as negative evidence.

\paragraph{Binding retention and conflict handling.}
A current binding is eligible for retention when its observation remains present and its local score exceeds a tracking threshold $\tau_{\mathrm{keep}}$. It is retained only if no other target's existing binding meets the same retention condition for that observation. Such contested bindings enter reassignment rather than being resolved by target iteration order; a newly scored candidate alone does not count as a competing retained binding. Targets without a retained binding activate memory assistance. Candidate pairs must exceed an association threshold $\tau_{\mathrm{assoc}}$. When entering recovery after losing a binding, they must also satisfy a prior-support threshold $\tau_{\mathrm{prior}}$. A target's verified initialization observation takes precedence when initialization occurs.

Eligible pairs are sorted by association score. Each pair is selected only if both its target and observation remain unassigned; ties are resolved by target index and then local track index. After allocation, pairs whose boxes exceed an overlap threshold $\tau_{\mathrm{IoU}}$ cannot both remain: verified initialization pairs have priority, followed by retained bindings and then association score. A suppressed pair does not trigger another search in the same frame. These operations generate provisional associations, not permission to train on the selected observations. They do not impose universal target- and track-margin rejection thresholds.

\subsection{Long-Term Distribution Details}\label{app:long_memory}

\paragraph{Initialization and bootstrap.}
Each local distribution is seeded with the normalized global features of initialization crops available at that view. New prototypes start with diagonal variance $\sigma_{\mathrm{init}}^2$ within the clipping interval $[\sigma_{\min}^2,\sigma_{\max}^2]$. Before the total count reaches $N_{\min}$, lack of a mature distribution does not veto an association. Bootstrap observations must instead satisfy prior support, association confidence, competing-target and competing-observation margins, and consecutive binding stability. They may populate the long-term distribution but are not treated as verified adaptation samples. Thus, bootstrap admission establishes the reference needed for verification without authorizing ordinary pseudo-label training before that reference is mature.

\paragraph{Admission conditions.}
Both margins are computed from a pre-allocation score matrix $Q_t^{(v)}(m,i)=s_{\mathrm{assoc}}^{(m,v)}(o_{t,i}^{(v)})$. Its rows include all targets already initialized in view $v$, including targets with retained bindings, and its columns include all current observations. Retention status and memory-assistance weights are determined first; all pair scores are then computed before acceptance filtering, greedy allocation, overlap suppression, and long-term verification. For a target $m$ and observation $o_i$, the margins are
\begin{equation}
\begin{aligned}
\Delta_{\mathrm{tgt}}(m,i)
&=Q_t^{(v)}(m,i)-\max_{m'\ne m}Q_t^{(v)}(m',i),\\
\Delta_{\mathrm{obs}}(m,i)
&=Q_t^{(v)}(m,i)-\max_{i'\ne i}Q_t^{(v)}(m,i').
\end{aligned}
\end{equation}
If a competitor set is empty, that margin imposes no restriction. The matrix is not reduced after assignment, so selecting one target--observation pair does not erase its competitors from the admission test. Long-memory admission requires association confidence at least $\tau_a$, both margins at least $\tau_\Delta$, and the same target--track binding for at least $\tau_T$ consecutive observed frames. Missing observations or a changed binding reset this count. Bootstrap additionally requires prior support at least $\tau_{\mathrm{boot}}$; after maturity, admission instead requires passing long-term verification. These are write conditions, not universal requirements for emitting a current association.

\paragraph{Prototype updates.}
An admitted canonical global feature $x$ is assigned to the prototype with maximum cosine similarity. If this similarity is below $\tau_{\mathrm{new}}$ and capacity remains, a new prototype is initialized with mean $x$, variance $\sigma_{\mathrm{init}}^2$, and count one. Otherwise the nearest prototype is updated without replacing another mode. After bank maturity, a new prototype can be created only from evidence that already passes the existing verifier. Maturity is assessed at bank level: a newly created prototype participates in matching even before its individual count reaches $N_{\min}$. This permits additional modes within the accepted region but does not ensure recovery from abrupt appearance changes outside it. Let $\mu_{\mathrm{old}}$ and $\sigma_{\mathrm{old}}^2$ be its parameters before the update and $\Delta=x-\mu_{\mathrm{old}}$. The update is
\begin{equation}
\begin{aligned}
\mu_{\mathrm{new}}
&=(1-\alpha)\mu_{\mathrm{old}}+\alpha x,\\
\sigma_{\mathrm{new}}^2
&=\operatorname{clip}_{[\sigma_{\min}^2,\sigma_{\max}^2]}
\left((1-\alpha)\sigma_{\mathrm{old}}^2+
\alpha(1-\alpha)\Delta^{\odot2}\right).
\end{aligned}
\label{eq:proto_update}
\end{equation}
The factor $(1-\alpha)$ in the residual term accounts for the moving mean. The prototype count is incremented by one for each newly admitted sample; it records accumulated support rather than the effective sample size of the exponential moving average. Variance clipping prevents degenerate uncertainty estimates. Matching uses minimum Mahalanobis distance, whereas update routing uses maximum cosine similarity.

\paragraph{Cross-view evidence.}
Each process maintains its own target distributions. A received long-memory event contains the canonical feature, source view, timestamp, and an observation identifier for deduplication. Sender-side admission does not automatically authorize a receiver-side update. At a mature receiver, an event may update the distribution only when the receiver has a local association for the same target at the current step that satisfies its own verification, confidence, margin, and stability conditions. The received feature must independently pass the receiver's pre-update verifier and meet the temporal eligibility rule in Appendix~\ref{app:association}. Stability and margins are measured on the receiver's local association, not inferred from the remote track identifier. Before maturity, the receiver instead requires a local association satisfying its bootstrap conditions, and the received feature must independently satisfy the bootstrap prior-support threshold in the canonical space. All checks use the receiver's statistics at the start of the step. Without qualifying local evidence, remote samples can support candidate association but cannot update the receiver's long-term distribution. Thus, cross-view evidence cannot directly mature that distribution while the target is unobserved locally. Repeated events are not counted twice. Local distributions can consequently differ while remaining comparable in the common canonical space.

\paragraph{Verification outcomes.}
A mature bank returns pass or fail according to the consistency threshold; an immature bank returns unavailable. Failed new or recovered proposals are removed without searching for a replacement in the same frame. A retained binding can remain visible in the output despite a failed check, but this does not admit its observation to long memory, Ridge training, or adapter training. An unavailable result does not by itself authorize learning; initialization labels and the restricted bootstrap procedure provide the starting evidence. All checks use pre-update statistics, so an observation cannot first change the reference against which it is verified.

\subsection{Online Adaptation Details}\label{app:adaptation}

\paragraph{Residual architecture.}
For each target--view pair and cue $k$, the residual branch has two linear layers with a ReLU nonlinearity:
\begin{equation}
h_{\phi_k}(g)=W_{2,k}\operatorname{ReLU}(W_{1,k}g+b_{1,k})+b_{2,k}.
\end{equation}
The input and output dimensions equal the canonical cue dimension $D_k$, and the hidden width is $\max(1,\lfloor D_k/4\rfloor)$. The first layer uses a standard variance-scaled initialization; the second-layer weights and bias are zero-initialized. Thus the residual output is initially zero without making both layers identically zero. The output is normalized before similarity computation, classifier fitting, and distribution anchoring. The canonical encoder, visibility estimator, and orientation predictor remain frozen. Adapters are local to each target--view pair and their parameters are not exchanged.

\paragraph{Sample permissions and negatives.}
Initialization observations carry their supplied labels. Other positive samples require admission under Sec.~\ref{sec:updates}, including receiver-side checks for cross-view samples. Training eligibility is stored separately from presence in a short-term proposal bank. A negative for target $m$ must carry an admitted identity $m'\ne m$; not being selected for $m$ is insufficient. Observations whose identity remains unresolved after conflict or overlap handling are excluded. Among eligible negatives, samples most similar to the anchor under the current target-$m$ adapter are selected for hard-negative training. This procedure does not treat an arbitrary co-present detection as a negative. When no eligible negative is available, the contrastive term is skipped rather than assigning an unverified negative label.

\paragraph{Cue aggregation and triplet construction.}
The positive pool combines training-eligible observations in the short-term support bank with the Ridge positive buffer, deduplicated by source view and observation identifier. Tentative support without training permission is excluded. Within each same-view or cross-view allocation, valid anchor--positive pairs are sampled uniformly without replacement, subject to the time-separation requirement. An observation cannot serve as both anchor and positive in a pair. Negatives follow the admitted hard-negative rule above.

All features in a triplet are recomputed from their cached canonical features through the same local adapter. The cue set is
\begin{equation}
\mathcal K(a,p,n)=\{0\}\cup
\{k\in\{1,\ldots,K\}:w_k(a)>0,\ w_k(p)>0,\ w_k(n)>0\}.
\end{equation}
Missing part masks exclude the corresponding parts. The triplet loss averages cue-level losses:
\begin{equation}
\ell_{\mathrm{tri}}(a,p,n)=
\frac{1}{|\mathcal K(a,p,n)|}
\sum_{k\in\mathcal K(a,p,n)}\ell_{\mathrm{tri},k}(a,p,n).
\end{equation}
For the receiver view $v$, same-view triplets use a local anchor and a distinct local same-target positive; cross-view triplets use a local anchor and an admitted same-target positive from another view. Negatives may come from either source if their receiver-side training permission is valid. The losses are
\begin{equation}
\mathcal L_{\mathrm{sv}}=
\frac{1}{|\mathcal T_{\mathrm{sv}}|}\sum_{\mathcal T_{\mathrm{sv}}}\ell_{\mathrm{tri}}(a,p,n),
\qquad
\mathcal L_{\mathrm{cv}}=
\frac{1}{|\mathcal T_{\mathrm{cv}}|}\sum_{\mathcal T_{\mathrm{cv}}}\ell_{\mathrm{tri}}(a,p,n).
\label{eq:memory_contrastive_loss}
\end{equation}
An empty triplet set contributes zero. The two sets are averaged independently before applying $\gamma_{\mathrm{cv}}$, so sample counts do not implicitly determine the relative loss weight.

\paragraph{Anchor eligibility and update order.}
The set $\mathcal P_+$ comprises distinct admitted target observations used as anchors or positives in the batch; negative observations are never pulled toward the target's Gaussian distribution. If no valid triplet exists, no contrastive or anchor-only adapter step is performed. Adapter optimization begins only when the receiver's target distribution is mature. The distribution snapshot taken before the current step's memory writes is held fixed throughout all optimizer steps for that batch. The optimizer differentiates the anchor loss through the adaptive global feature but not through Gaussian statistics or the canonical encoder. Part features receive the contrastive loss without a separate Gaussian penalty.

After optimizing the adapter, the receiver invalidates adaptive-feature caches and recomputes them from stored canonical features. Orientation/cue Ridge classifiers with both admitted classes are refit in the new adaptive space using Appendix~\ref{app:association}; missing-class models are disabled, and old weights are never used with new features. Buffer changes also trigger refitting without an adapter update. Available classifiers are aggregated; if none applies, scoring uses the prior fallback. This refresh is completed before the next association step. No backbone forward pass on historical images and no synchronization of adapter parameters are required.

\section{Experimental Details}\label{app:experimental_details}
\subsection{Implementation Settings}\label{app:configuration}

The canonical encoder uses the trained ResNet-18 checkpoint from OCLReID~\cite{ye2024oclreid}.

Tables~\ref{tab:configuration_association} and~\ref{tab:configuration_adaptation} summarize the configuration of the complete model. Ablation-specific changes are described in Appendix~\ref{app:ablation_controls}. The canonical encoder and its feature output path, visibility estimator, and orientation predictor remain frozen; only the local residual adapters are optimized. Their architecture and initialization are specified above. Offline references are trained using two NVIDIA RTX A6000 GPUs, while evaluation uses a single RTX A6000 GPU.

\begin{table}[htbp]
\centering
\caption{Association, verification, and admission settings. Capacities count observations or prototypes, not individual feature coordinates.}
\label{tab:configuration_association}
\small
\begin{tabular}{p{0.69\linewidth}p{0.23\linewidth}}
\toprule
Setting & Value \\
\midrule
Detector / canonical backbone & YOLOX-L / ResNet-18 \\
Canonical checkpoint source & OCLReID-trained ResNet-18 \\
Number of part cues $K$ & 4 \\
Global and part dimensions $D_k$ & 512 \\
Detection score / NMS IoU threshold & 0.60 / 0.35 \\
Support aggregation $L$ & 3 \\
Active memory fusion weight $\lambda_{\mathrm{mem}}$ & 0.80 \\
Prior global / mean-part weights & 0.70 / 0.30 \\
Part-support threshold $\tau_p$ / margin $\delta_p$ & 0.80 / 0.005 \\
Binding retention $\tau_{\mathrm{keep}}$ & 0.60 \\
Association acceptance $\tau_{\mathrm{assoc}}$ & 0.70 \\
Recovery prior support $\tau_{\mathrm{prior}}$ & 0.72 \\
Assignment overlap threshold $\tau_{\mathrm{IoU}}$ & 0.70 \\
Maximum remote-observation age $\Delta_{\mathrm{sync}}$ & 0.10 s \\
Short support capacity per target and source view & 64 observations \\
\midrule
Gaussian capacity per target--view pair & 4 prototypes \\
Prototype update rate $\alpha$ & 0.05 \\
Initial diagonal variance $\sigma_{\mathrm{init}}^2$ & $1/512$ \\
Variance bounds $[\sigma_{\min}^2,\sigma_{\max}^2]$ & $[10^{-3},1]$ \\
Distance denominator stabilizer $\epsilon$ & $10^{-6}$ \\
New-prototype cosine threshold $\tau_{\mathrm{new}}$ & 0.75 \\
Bank maturity $N_{\min}$ & 10 observations \\
Verification score threshold $\tau_\ell$ & 0.50 \\
Admission score $\tau_a$ / competing margins $\tau_\Delta$ & 0.90 / 0.15 \\
Consecutive binding stability $\tau_T$ & 10 observed frames \\
Bootstrap prior-support threshold $\tau_{\mathrm{boot}}$ & 0.80 \\
\bottomrule
\end{tabular}
\end{table}

\begin{table}[htbp]
\centering
\caption{Local adapter learning and classifier settings. Ridge regularization refers to the class-normalized objective in Eq.~\ref{eq:ridge_fit}.}
\label{tab:configuration_adaptation}
\small
\begin{tabular}{p{0.69\linewidth}p{0.23\linewidth}}
\toprule
Setting & Value \\
\midrule
Adapter hidden width & 128 \\
Optimizer & Adam \\
Learning rate / weight decay & $10^{-4}$ / $10^{-4}$ \\
Optimization interval per target--view pair & 5 newly admitted local observations \\
Optimizer steps per eligible batch & 1 \\
Maximum / minimum triplets per batch & 16 / 4 \\
Same-view / cross-view sampling proportion & 0.5 / 0.5 \\
Minimum same-view positive time separation & 0.10 s \\
Triplet margin & 0.20 \\
Cross-view loss weight $\gamma_{\mathrm{cv}}$ & 0.50 \\
Distribution-anchor weight $\beta_{\mathrm{dist}}$ & 0.004 \\
Ridge regularization $\lambda_R$ & 0.01 \\
Ridge buffer per target--view pair and class & 32 observations \\
Contrastive negative pool per target--view pair & 256 observations \\
\bottomrule
\end{tabular}
\end{table}

\paragraph{Buffer and update conventions.}
Initialization references are retained separately from the bounded short support bank. Ridge buffers use FIFO eviction and retain the visibility information of each observation. Their capacity is per class, not per cue; each orientation/cue classifier uses the applicable subset. Contrastive negatives must have an admitted identity different from the target. Remote observations are used only after receipt, with timestamps no later than the current processing time and ages within $\Delta_{\mathrm{sync}}$.

The optimization interval counts newly admitted local observations with training permission, not elapsed video frames or received messages. Each eligible batch contains at least four valid triplets. The sampling proportions specify the intended allocation; unavailable triplets are not duplicated to fill a batch. The two loss groups are averaged separately as in Eq.~\ref{eq:memory_contrastive_loss}. Long-memory writes and pseudo-label training use the same confidence, margin, and stability requirements after bank maturity. Bootstrap observations only establish the reference and do not count toward adapter-training eligibility. Adapter updates use pre-write Gaussian statistics, followed by feature remapping and Ridge refitting before the next association step.

\subsection{Evaluation Protocol and Baseline Adaptation}\label{app:evaluation}

\paragraph{Target selection and initialization.}
Each scene contains two designated target identities, with a fixed mapping to dataset identities shared across views and methods. M3Track reads this mapping from the target metadata; MvMHAT uses dataset identities 1 and 2. The four online baselines receive a bounding box at each target's first annotated occurrence in each view's test segment. Visible target frames remain in evaluation even if initialization fails. Only selected identities are retained in the evaluation ground truth; missed targets and false outputs remain subject to the evaluator's matching rules.

\paragraph{Metric computation and aggregation.}
Tracking metrics are computed with TrackEval, treating each view as a separate sequence and retaining selected-target GT. The selected-target exporter uses the initialized target label, mapped to its dataset ID, as the prediction ID rather than the underlying local track ID. This fixes exported labels but does not constrain metric matching by label equality: CLEAR uses spatial matching with its temporal preference, and Identity uses global trajectory assignment. Dataset-level results combine sequence statistics rather than averaging scene scores; IDs are summed, and MOTP measures matched-box IoU. Cross-view scores are computed for synchronized view pairs, then averaged over scenes on M3Track and over all scene--view pairs on MvMHAT. Thus, four-view scenes contribute six pairs and three-view scenes three pairs. Both benchmark wrappers use the same pairwise evaluation function described below. These metrics assess tracking and association consistency rather than enforcing the initialized identity name at every match.

\paragraph{Identity switches.}
The IDs columns use the CLEAR IDSW field returned by TrackEval, also used in MOTA. An identity switch is counted when a GT identity is matched to a prediction ID different from its previous matched prediction ID. A gap does not by itself reset that previous ID. This is not a count of wrong-identity frames or the separate selected-target diagnostic switch counter.

\paragraph{Cross-view matching and empty frames.}
Within each view, each GT box is matched to the prediction with the
highest IoU among candidates with IoU at least 0.5; if no such prediction
exists, the GT is treated as unmatched. The selected prediction label is
then used to construct the GT-to-prediction identity map. A shared GT identity is cross-view correct when its matched prediction labels agree across the two views; equality to the GT identity label is not required.

Let $G_t$ be the number of GT identities common to both views, $R_t$ the number of distinct positive prediction labels shared by the two GT-to-prediction maps, and $C_t$ the number of cross-view-correct GT identities. Let $Q_t$ count GT entries in view $a$ whose mapped prediction label occurs in both maps. Frame precision and recall are $C_t/Q_t$ and $C_t/G_t$, respectively. Frame MDA is
\[
\frac{C_t}{G_t+\max(G_t-C_t,0)+\max(R_t-C_t,0)}.
\]
Zero-denominator values contribute zero. If either view has no GT, frame precision, recall, and MDA also contribute zero; such frames remain in the averaging denominator. With GT in both views but no predictions in one view, $C_t=R_t=Q_t=0$, so these three frame scores are zero. CVIDP, CVIDR, and MDA are frame means; pair CVIDF1 is the harmonic mean of pair CVIDP and CVIDR before averaging across pairs.

For CVMA, let $E_t=\max(R_t-C_t,0)$ and $M_t=2\max(G_t-R_t,0)$. Let $F_t$ count distinct predicted labels present in both views but absent from one view's GT-to-prediction map, summed over the two directions. With $N_t$ the total number of GT boxes in both views,
\[
\mathrm{CVMA}=1-\frac{\sum_t(2E_t+F_t+M_t)}{\sum_t N_t}.
\]
When either view has no GT, $E_t,F_t,M_t$ are zero, but available GT boxes still contribute to $N_t$. A zero total GT-box count gives CVMA zero.

\paragraph{Baseline adaptation.}
The online baselines run independently for each target and view,
without exchanging observations, features, trajectories, or
association decisions across cameras. Since these baselines do
not perform cross-view association, we assign outputs for the
same initialized target a common label across views. This
supplies cross-view label correspondence without requiring the
baselines to infer it. Their cross-view scores are then computed
from these aligned outputs using the same evaluation rules.
\subsection{Full-Model Ablation Controls}\label{app:ablation_controls}

The full-model controls retain the same observations, target initialization, association procedure, and non-ablated settings. They differ from the cumulative component rows because each changes one specified aspect of Full.
Reliability control denotes the non-Gaussian admission conditions: the confidence threshold, both competing margins, and consecutive binding stability. It is distinct from V-Upd., which makes long-term verification a prerequisite for persistent writes and training. Base disables these additional reliability conditions, cross-view assistance, adapter optimization, and Gaussian verification; it retains local association and classifier fitting. Detection filtering, association acceptance, one-to-one allocation, and overlap handling remain common to all variants.

\paragraph{Without reliability control.}
Full w/o Rel. removes the confidence, competing-margin, and binding-stability conditions from ordinary memory and training admission. It retains Gaussian verification and verification-gated writes, cross-view assistance, and adapter learning. Initialization, restricted bootstrap, bank maturity, and valid-sample requirements remain unchanged. Thus, this control removes non-Gaussian reliability screening rather than long-term verification or basic association safeguards.

\paragraph{Cumulative online adaptation.}
For the cumulative Online Contrastive variant without Gauss., training admission uses the remaining reliability conditions without requiring long-term verification or a mature Gaussian bank, and the distribution-anchor loss is omitted. Thus, adding Gauss. changes both verification and its associated learning constraints.

\paragraph{Without online adaptation.}
The local adapters remain at their identity initialization, and no contrastive or distribution-anchor optimization is performed. Cross-view assistance, canonical-space verification, and memory admission remain active. Ridge classifiers are still refit when their admitted training buffers change, so this control removes representation learning rather than all online state updates.

\paragraph{Without verified updates.}
Long-term verification remains available for association decisions, but its pass/fail outcome no longer gates persistent-memory admission or training eligibility for the adapters and Ridge classifiers. This removal applies to both local and received observations. Confidence, competing margins, binding stability, and the requirement for a locally admitted same-target observation at the receiver are retained. Bootstrap and the requirement for a mature anchor before adapter learning are unchanged. This control tests verification-dependent write permissions rather than removing the entire reliability controller.

\paragraph{Adaptive verification features.}
This control replaces only the normalized canonical global query in long-term verification with its normalized output from the current local adapter; Gaussian statistics remain in canonical space, and their update rule, write permissions, and adaptation losses are unchanged. Both variants use the same verification threshold $\tau_\ell=0.50$. This comparison tests query substitution at a common threshold, not matched acceptance rates or the necessity of the entire two-space design.

\clearpage

\subsection{Additional Qualitative Analysis}\label{app:qualitative}

\begin{figure}[htbp]
\centering
\includegraphics[width=\linewidth]{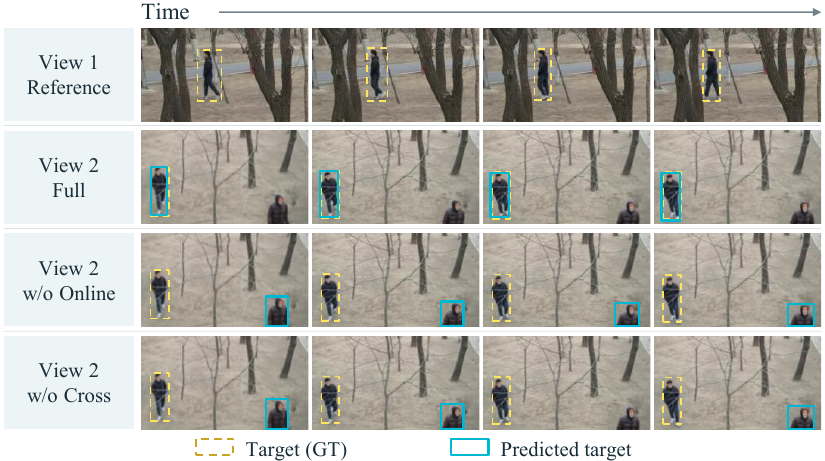}
\caption{Additional comparison in the M3Track wood scene. Columns correspond to dataset frames 1861, 1867, 1871, and 1875. The reference view is shown once; the other rows use identical images and crops for the three configurations. Yellow dashed boxes denote the target and cyan boxes denote confirmed outputs; tentative candidates are not drawn. Displayed View~1 and View~2 correspond to source view2 and view1, respectively.}
\label{fig:qualitative_wood}
\end{figure}

Figure~\ref{fig:qualitative_wood} shows an additional identity maintenance example in the M3Track wood scene. The selected target and a nearby pedestrian have similar dark clothing. The full model maintains the target identity throughout the displayed interval, whereas the configuration without online contrastive adaptation assigns it to the other pedestrian. Without cross-view assistance, the controller also selects the distractor before and after a brief interval with no confirmed output. Thus, the two ablations differ in whether an incorrect candidate is emitted, rather than exhibiting two consistently distinct failure modes.

The absence of a cyan box in the third column of the last row corresponds to this interruption. It lasts seven consecutive dataset frames (1868--1874), during which an incorrect tentative candidate remains available internally. The interruption therefore does not indicate that the identity has been corrected. These examples illustrate tracking behavior rather than isolate the effects of representation spaces or verification-dependent update permissions.

\end{document}